\documentclass[11pt,letterpaper]{article}
\usepackage[utf8]{inputenc}
\usepackage{amsmath}
\usepackage{amsfonts}
\usepackage{amssymb}
\usepackage[margin=2.5cm]{geometry}
\usepackage{setspace}
\setdisplayskipstretch{0.3}
\usepackage{graphicx, caption}
\usepackage{subcaption}
\usepackage{wrapfig}
\graphicspath{ {images/} }
\usepackage[export]{adjustbox}
\usepackage{microtype}
\DisableLigatures{encoding = *, family = * }
\begin{document}

\begin{center}
\begin{large}
PREPRINT

\textbf{The Location of Optimal Object Colors with More Than Two Transitions}
\end{large}

Scott A. Burns, University of Illinois at Urbana-Champaign, Urbana, IL 61801.

\vspace{-0.5cm}
Email: scottallenburns@gmail.com

\vspace{-0.5cm}
May 14, 2021
\end{center}

\textit{\textbf{Preprint Status Notice}: This preprint contains substantially less material than the version that was eventually accepted for publication in Color Research and Application. In particular, most of the material contained in the online supplementary documentation (Reference 27) as been incorporated into the final version at the request of the reviewers. In addition, an analysis was performed on the more modern ``physiologically-relevant" color-matching functions that are transformed from cone fundamentals. This additional analysis is included at the end of the final peer-reviewed version.}

\textit{Abstract: The chromaticity diagram associated with the 1931 $2^{\circ}$ CIE color matching functions is shown to be slightly non-convex. While having no impact on practical colorimetric computations, the non-convexity does have a significant impact on the shape of some optimal object color reflectance distributions associated with the outer surface of the object color solid. Instead of the usual two-transition Schrödinger form, many optimal colors exhibit higher transition counts. A linear programming formulation is developed and is used to locate where these higher-transition optimal object colors reside on the object color solid surface. The regions of higher transition count appear to have a point-symmetric complementary structure.}

%=======================================================================

\textbf{1. Introduction}

The color-matching functions (CMFs), when plotted as a set of three-dimensional vectors in tristimulus space, form an origin-based cone enveloping all possible colors. Nested within this cone is the object color solid (OCS). It contains all ``object colors" that can be produced by a non-fluorescent reflecting surface (i.e., having a spectral reflectance distribution between zero and one), under the action of some specified illuminant. The outer surface of the OCS contains the ``optimal object colors." Historically, the optimal colors are believed to be associated with rectangular spectral reflectance distributions having only values of zero and one and at most two sudden transitions, known as Schrödinger colors.$^{1}$ This belief has been echoed in color science publications for many decades.$^{2-8}$

West and Brill showed that one condition for all optimal colors to have the Schrödinger form is that the chromaticity diagram (the so-called horseshoe diagram) be convex.$^{9}$ Most historic proofs of optimal colors having the Schrödinger form assume this convexity, either explicitly or implicitly.

Recently, Davis demonstrated that the CIE 1931 CMFs produce a non-convex chromaticity diagram.$^{10}$ His presentation is highly mathematical and intended for a very specialized audience. The first part of this paper will demonstrate this non-convexity in a simpler way, using only a single MATLAB function.

The non-convexity is very slight and has no impact on practical colorimetric computations. It does, however, have a significant impact on the shape of some optimal object color reflectance distributions. Instead of the usual two-transition Schrödinger form, many optimal colors exhibit higher transition counts. A linear programming approach will be developed and be used to compute the reflectance distribution of optimal object colors, located anywhere on the OCS surface. It will be used to map regions where these higher-transition optimal colors reside on the OCS surface.

%=======================================================================

\textbf{2. Convexity of the Chromaticity Diagram}

Here we are considering the 1931 $2^{\circ}$ CIE Standard Colorimetric Observer Data (CMFs) that span 360 nm to 830 nm in 1 nm intervals.$^{11,12}$ We designate this 471$\times$3 matrix with the symbol $A$. The chromaticity coordinates associated with $A$ are found by dividing each row of $A$ by the row sum. The first two columns are then identified as $x$ and $y$ chromaticity coordinates.  In MATLAB this is accomplished with the statement $\texttt{xy = A(:,1:2)./sum(A,2);}$, where $xy$ is a 471$\times$2 matrix of chromaticity coordinates, producing the well-known ``horseshoe" chromaticity diagram.

We now use the notion of a convex hull to assess the convexity of the chromaticity diagram. When applied to a discrete set of points in two dimensions, the convex hull is the smallest convex polygon that contains all of the points. If it turns out that any of the chromaticity coordinates of the horseshoe reside in the interior of the convex hull, then the horseshoe is not convex.

The convex hull is generated in MATLAB with the statement $\texttt{k = convhull(xy)}$. The vector $k$ contains the indices of rows of $xy$ that form the convex hull boundary. It turns out that only 161 of the 471 rows of $xy$ contribute to this boundary. Figure 1 depicts graphically where they are. The black dots are those that form the convex hull boundary and the red dots are those that fall inside of it.

\begin{figure}[p]
\includegraphics[width=1\textwidth]{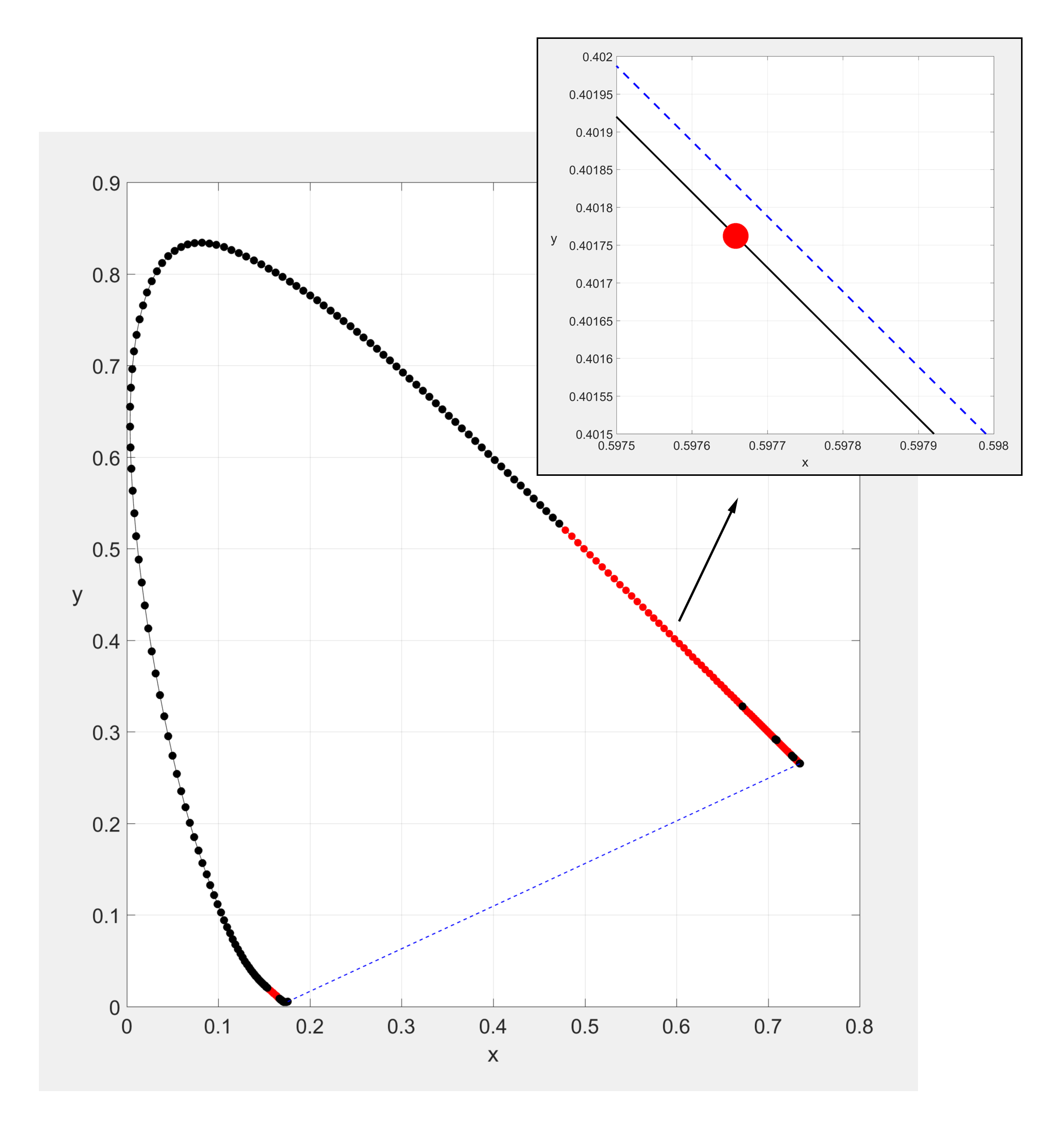}
\centering
\caption{Chromaticity diagram (black solid line) and convex hull enveloping it (blue dashed line). Red dots are points on chromaticity diagram that fall slightly inside the convex hull, indicating a slight non-convexity of the horseshoe for portions above 574 nm and also a small portion running 435 nm to 453 nm.}
\end{figure}

Close examination of one of the red dots (figure inset) shows that the horseshoe polygon (solid black line) falls slightly inside the convex hull polygon (blue dashed line) in some areas. There are two main regions of non-convexity: several large pockets of wavelengths above 574 nm, and another segment running 435 nm to 453 nm. There are also many pockets of non-convexity at both extreme ends of the visible range.

The non-convexity persists with lower-resolution versions of the 1 nm CMFs. Figure 2 shows the non-convexity of 5 nm and 10 nm versions of CMFs that are commonly used in colorimetric computations. It is evident that the non-convexity is present here as well.

\begin{figure}[h]
\captionsetup{width=0.95\linewidth}
\includegraphics[width=0.95\textwidth]{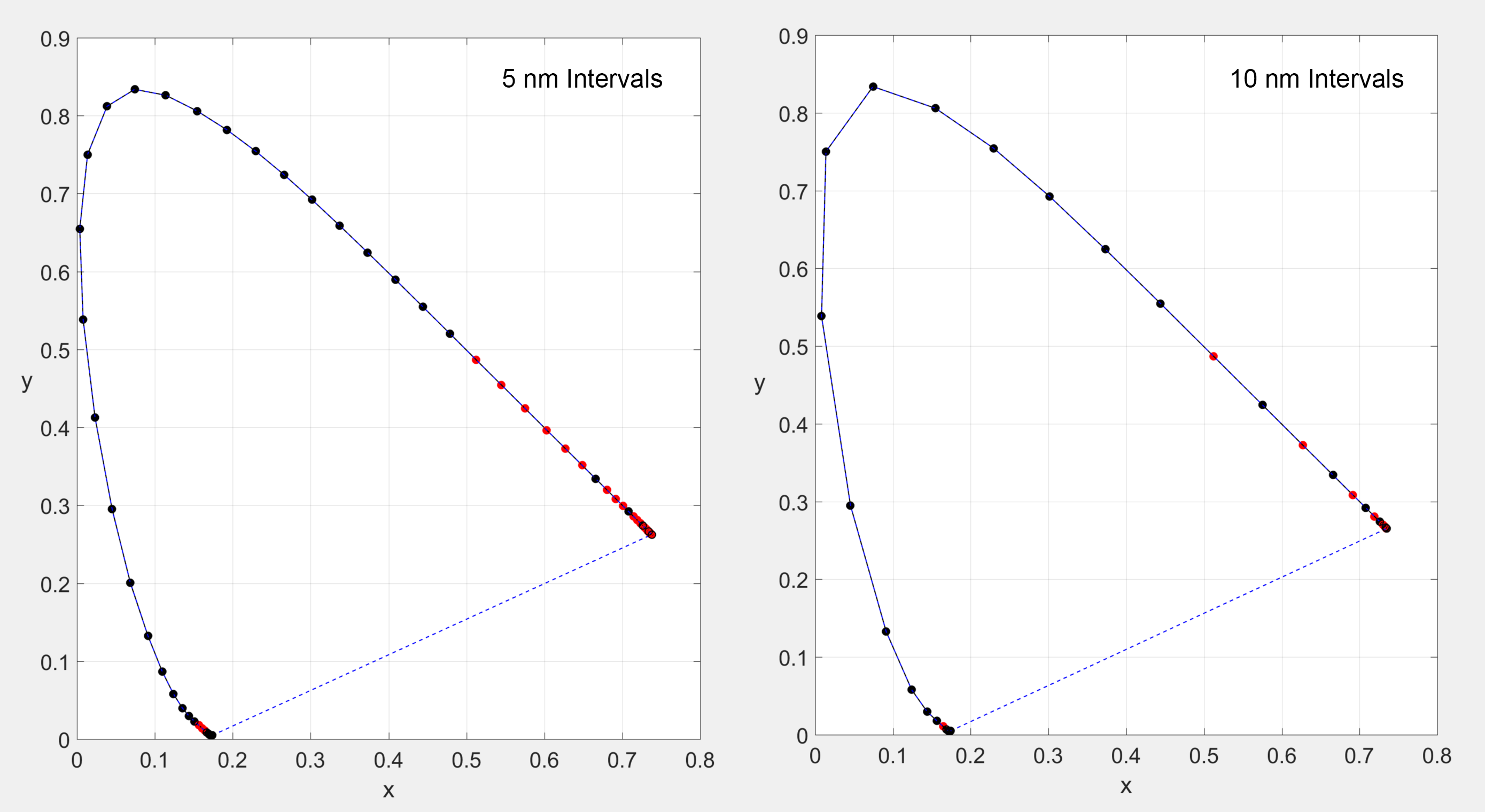}
\centering
\caption{Convex hull analysis of lower-resolution versions of the 1 nm CMFs, having 5 nm wavelength intervals (left) and 10 nm intervals (right). Red dots show chromaticity coordinates falling slightly inside the convex hull boundary.}
\end{figure}

This non-convexity is extremely small. In chromaticity space units, the horseshoe and convex hull polygons are only about 0.00005 units apart in the Figure 1 inset. This very small difference will have no practical impact on typical colorimetric calculations. It turns out, however, that it does have a significant impact on the shape of optimal object colors, as discussed in the next sections.

%=======================================================================

\textbf{3. Computing Optimal Object Color Reflectance Curves}

The standard approach to characterizing the object color solid is to start with two-transition Schrödinger reflectance curves and map them into tristimulus space to form the boundary of the solid.$^{2-5,13}$ An alternate approach is to make no assumption of the shape of the optimal reflectance curve, but seek points on the surface of the OCS by optimization methods. In this paper, we follow the latter approach using linear programming (LP). Historically, LP has been used in colorimetric calculations to compute metamer mismatch volumes,$^{14-17}$ to design illuminants to achieve desired rendering effects,$^{18}$ and to compute subtractive mixture recipes for colorants based on the Kubelka-Munk theory.$^{19-21}$

The idea of using LP to identify the OCS surface is not new. In his 1969 paper on limits of metamerism (i.e., metamer mismatch volumes), Allen mentions in passing (without any details) that LP could be used to compute MacAdam limits (optimal object colors).$^{14}$ Ohta and Wyszecki repeat Allen's claim in their 1975 paper on metamer mismatch volumes, but only go so far as to state that it would be achieved with a simplified version of some of the equations they present.$^{15}$ In 2010, Li et al explicitly use LP to determine the shape of cross-sections of the OCS by first identifying key locations on each cross section by LP, and then by using LP repeatedly to fill in the curves between these key locations.$^{22}$ Although their LP formulation directly computes the shape of the reflectance curves as a by-product of the computation, the focus of their paper is instead on identifying the shape of the OCS in tristimulus space. They apparently did not encounter (or perhaps notice) any cases where the two-transition assumption was violated (other than inevitable artifacts of discretization at the two transition locations, which they did mention).

The LP presented in this paper uses only one basic colorimetric equation:

\vspace{-0.8cm}
\begin{equation}
{XYZ}_W = {A_W}'\;\rho.
\end{equation}

In this equation, ${XYZ}_W$ is a $3{\times}1$ vector of illuminant-W-referenced tristimulus values, and $\rho$ is an $n{\times}1$ vector of reflectance values (0-1), where $n$ is the number of wavelength intervals used to discretize continuous spectral distributions over the visible range. Matrix $A_W$ is an $n{\times}3$ matrix of illuminant-W-referenced color matching functions, computed as $\overline{W} A$, where $\overline{W}$ is an $n{\times}n$ matrix with illuminant $W$ on the main diagonal and zeros elsewhere. Prime denotes matrix transpose. As mentioned in Section 2, matrix $A$ is an $n{\times}3$ array of color matching functions (CMFs) by columns, $A=[\overline{x}, \overline{y}, \overline{z}]$. It is assumed that the illuminant has been normalized so that the scalar product ${\overline{y}}\,'\cdot W$ equals 100. For the standard 1931 $2^{\circ}$ CIE standard CMFs with 1 nm resolution, $n$ is 471.

Note that the reflective/transmissive version of tristimulus values are being used here (as opposed to the emissive version), so all tristimulus values are referenced to an illuminant, hence the $W$ subscript. Even though different subscripts may be used to identify special tristimulus values in this paper, it is understood that all of them are $W$-referenced. For example, the white point, ${XYZ}_{wp}$ is the tristimulus triplet formed as the three row sums of ${A_W}'$ (three column sums of $A_W$).

Figure 3 presents a cut-away view of the interior of the OCS. Four key locations are identified, (1) the white point, ${XYZ}_{wp}$, (2) the 50\% gray halfway point between the origin and the white point, ${XYZ}_{\mathit{50\%}} = {XYZ}_{wp}/2$, (3) an arbitrary point within the OCS, ${XYZ}_{targ}$, which defines a direction with respect to ${XYZ}_{\mathit{50\%}}$, and (4) the point of intersection of a ray in that direction with the OCS outer surface, ${XYZ}_{opt}$.

\begin{figure}[!]
\captionsetup{width=0.95\linewidth}
\includegraphics[width=0.95\textwidth]{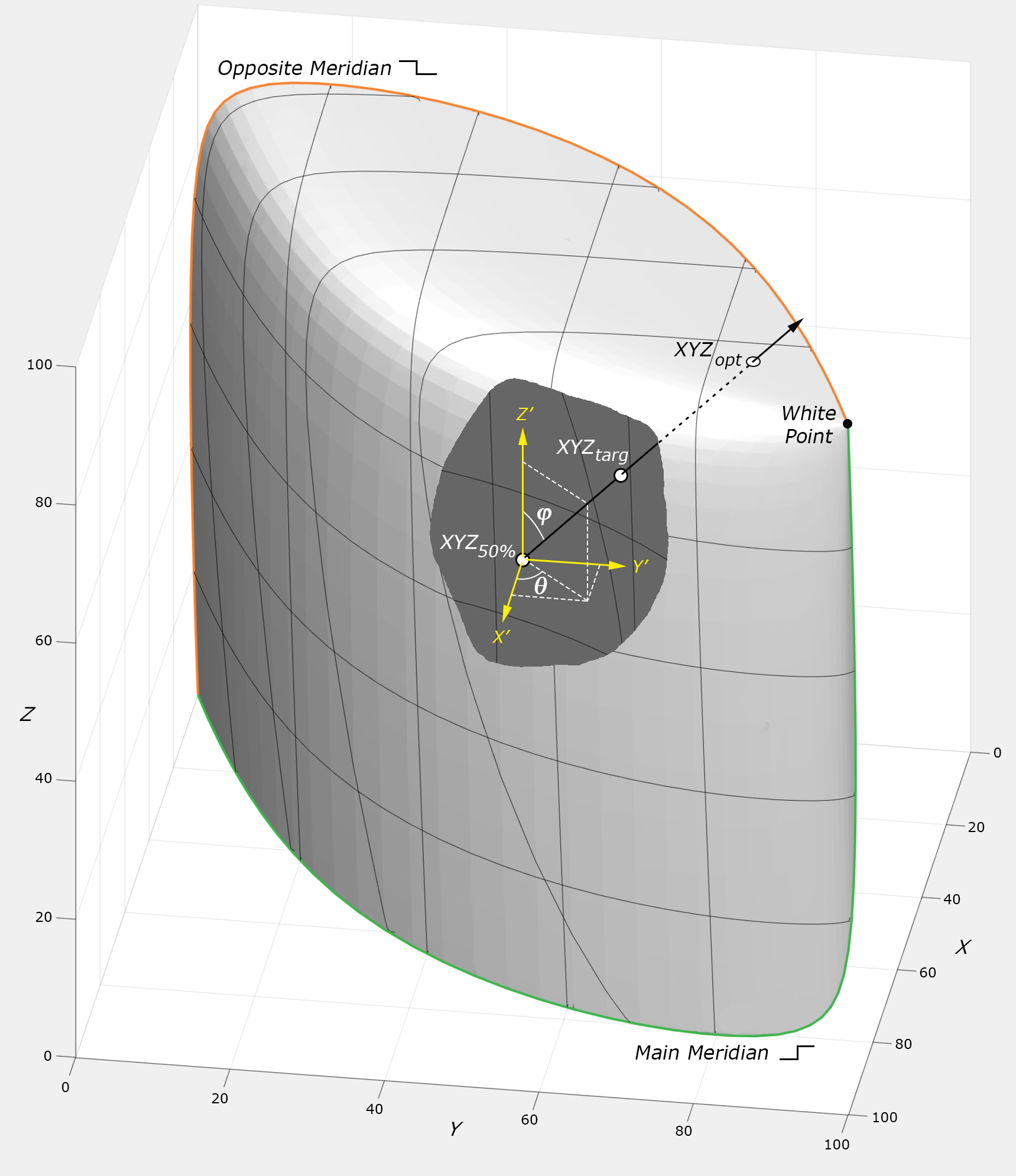}
\caption{An object color solid with key locations identified.}
\end{figure}

A spherical coordinate system is established, based at ${XYZ}_{\mathit{50\%}}$. Every ${XYZ}_{opt}$ on the surface is associated with a unique $(\theta,\varphi )$ pair. The spherical coordinate system will come in handy in a subsequent section, when it comes time to plot a dense array of optimal color data as viewed from the interior of the OCS.

Consider the vector ${XYZ}_{targ} - {XYZ}_{\mathit{50\%}}$. When this vector is based at ${XYZ}_{\mathit{50\%}}$, it points in the direction of the ray shown in Figure 3. Next, consider a scalar, $c$, which serves to magnify the vector and locate an arbitrary point, ${XYZ}$, along the ray:

\vspace{-0.8cm}
\begin{equation}
{XYZ} = {XYZ}_{\mathit{50\%}} + c ({XYZ}_{targ} - {XYZ}_{\mathit{50\%}}).
\end{equation}

Substituting for ${XYZ}$ using Equation 1, we now have an equation in terms of our unknowns, $\rho$ and $c$. We seek to maximize $c$ to move as far along the ray as possible, while constraining $\rho$ to be between 0 and 1:

\vspace{-0.8cm}
\begin{equation}
\begin{split}
\underset{\rho,c}{\mathsf{max}}\;\;&c\\
\mathsf{s.t.}\;\;&{A_W}'\;\rho = {XYZ}_{\mathit{50\%}} + c ({XYZ}_{targ} - {XYZ}_{\mathit{50\%}})\\
&0\leq \rho \leq 1.
\end{split}
\end{equation}

This is a linear program. To solve this LP with MATLAB's ``linprog" function, it needs to be recast in standard form:

\vspace{-0.8cm}
\begin{equation}
\begin{split}
\underset{x}{\mathsf{min}}\;\;&f' x\\
\mathsf{s.t.}\;\;&A_{eq} x = b_{eq}\\
&x_{lb} \leq x \leq x_{ub},
\end{split}
\end{equation}

where $f, x, b_{eq}, x_{lb}$, and $x_{ub}$ are all vectors and $A_{eq}$ is a matrix. This recasting of Equation 3 is accomplished by concatenating $\rho$ and $c$ into a single 472$\times$1 vector and by minimizing $-c$.

\vspace{-0.8cm}
\begin{equation}
\begin{split}
\underset{\rho,c}{\mathsf{min}}\;\;&-c\\
\mathsf{s.t.}\;\;
&\left[\begin{array}{c|c}
{A_W}\,' & {XYZ}_{\mathit{50\%}} - {XYZ}_{targ}
\end{array}\right]
\begin{Bmatrix} \rho \\ c \end{Bmatrix} = {XYZ}_{\mathit{50\%}}
\\
&\;0\leq \rho \leq 1.
\end{split}
\end{equation}

The following MATLAB statements produce the optimal color reflectance distribution, $\rho$, corresponding to a point on the surface of the OCS that is the intersection of a ray emanating from ${XYZ}_{\mathit{50\%}}$, passing through ${XYZ}_{targ}$:

\begin{quote}
$\texttt{ }\\
\texttt{>> options = optimoptions('linprog', 'Algorithm', 'dual-simplex',...}\\
\texttt{\qquad 'OptimalityTolerance', 1e-9, 'ConstraintTolerance', 3e-9);}\\
\texttt{>> x = linprog([zeros(471,1);-1], [\,] ,[\,], [Aw',XYZ50-XYZtarg], XYZ50,...}\\
\texttt{\qquad [zeros(471,1);-Inf], [ones(471,1);Inf], options);}\\
\texttt{>> rho = x(1:471);}$
\end{quote}

The two tolerances used internally by the LP solver are tightened up considerably. These values were determined experimentally to avoid premature termination of the LP. Note that the MATLAB Optimization Toolbox is required for function linprog.

The vast majority of reflectance curves generated by this LP will have the expected two-transition structure. For example, using a target of $\texttt{XYZtarg=[10;40;30]}$, the LP gives the reflectance shown on the left side of Figure 4. In some cases, however, more than two transitions are obtained. For example, if the target is changed to $\texttt{XYZtarg=[49.1;40.3;25.0]}$, a four-transition reflectance curve is produced, as shown on the right side of Figure 4.

\begin{figure}[h]
\captionsetup{width=1\linewidth}
\includegraphics[width=1\textwidth]{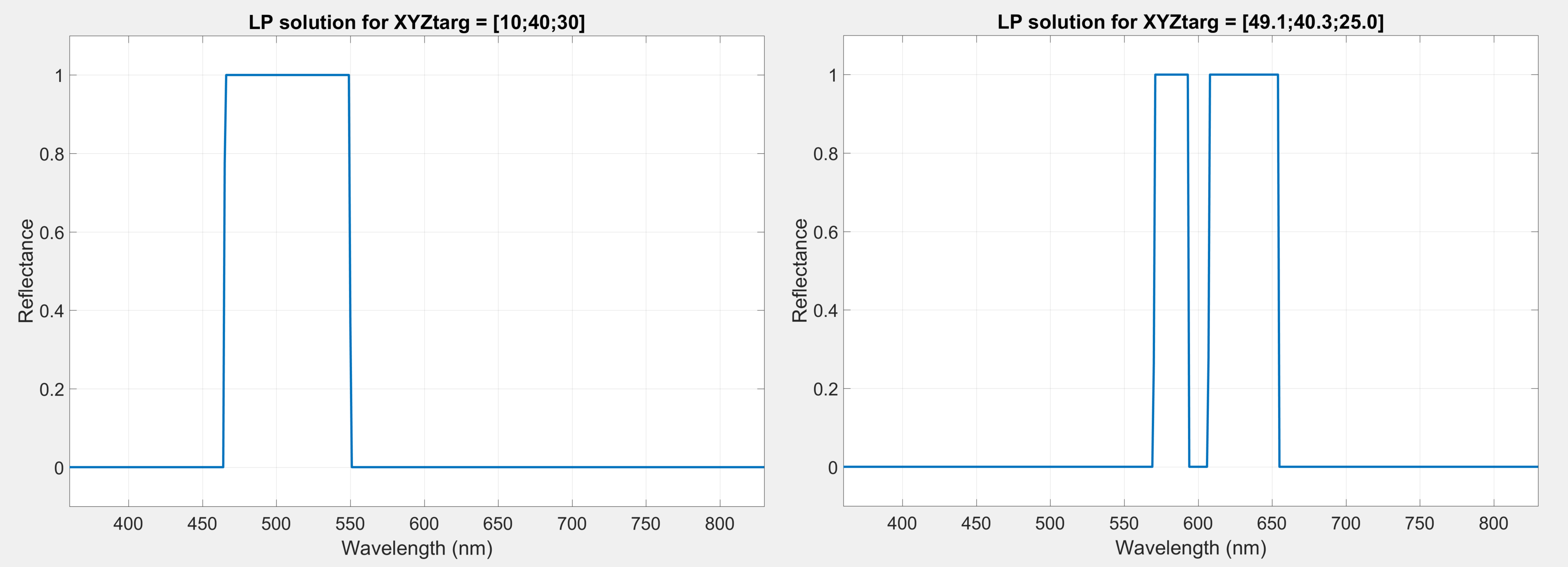}
\caption{Reflectance curves produced by the LP for two different targets.}
\end{figure}

It is natural to question the higher-transition result. Is it just an artifact of the LP solution process? Is it really on the outer surface of the OCS? In other words, is there a two-transition solution along the same ray that is even farther from ${XYZ}_{\mathit{50\%}}$, which would make the high-transition solution an interior point of the OCS, and not optimal? The LP solution process involves pivots into and out of an active basis, using tolerances to decide if optimality conditions have been numerically satisfied. Is it possible that the higher-transition solutions are just barely sub-optimal points that happen to satisfy the optimality conditions numerically within the specified tolerances? These issues will be addressed in subsequent sections, but first, a survey of how often and where the high-transition LP solutions occur will be presented.

%=======================================================================

\textbf{4. A Map of High-Transition LP Solutions}

Figure 3 presented a spherical coordinate system based at ${XYZ}_{\mathit{50\%}}$. A second rectangular coordinate system was also shown in that figure, defined as $(X', Y', Z') = (X, Y, Z) - {XYZ}_{\mathit{50\%}}$. Suppose we wish to visualize the OCS surface by placing our viewpoint at ${XYZ}_{\mathit{50\%}}$ and directing our view either up or down the $Z'$ axis. Figure 5 shows two polar plots of how we can arrange a mapping of the spherical coordinate system, in both cases, placing the $X'$ axis running to the right. The left plot in Figure 5 visualizes the upper half of the OCS and the right plot visualizes the lower half.

\begin{figure}[h]
\captionsetup{width=1\linewidth}
\includegraphics[width=1\textwidth]{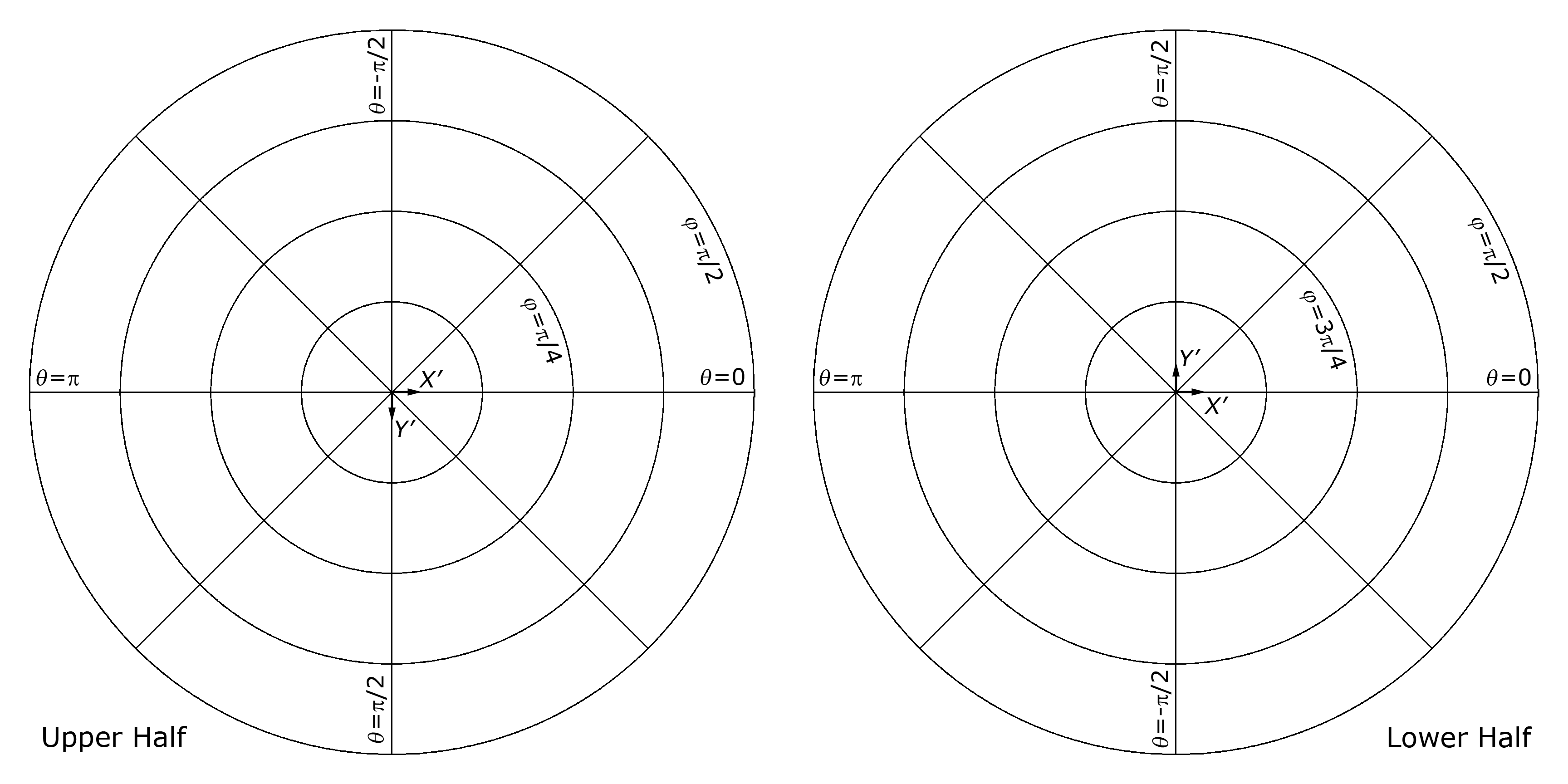}
\caption{Polar mapping for upper and lower halves of OCS.}
\end{figure}

\begin{figure}[p]
\captionsetup{width=0.7\linewidth}
\includegraphics[width=0.65\textwidth]{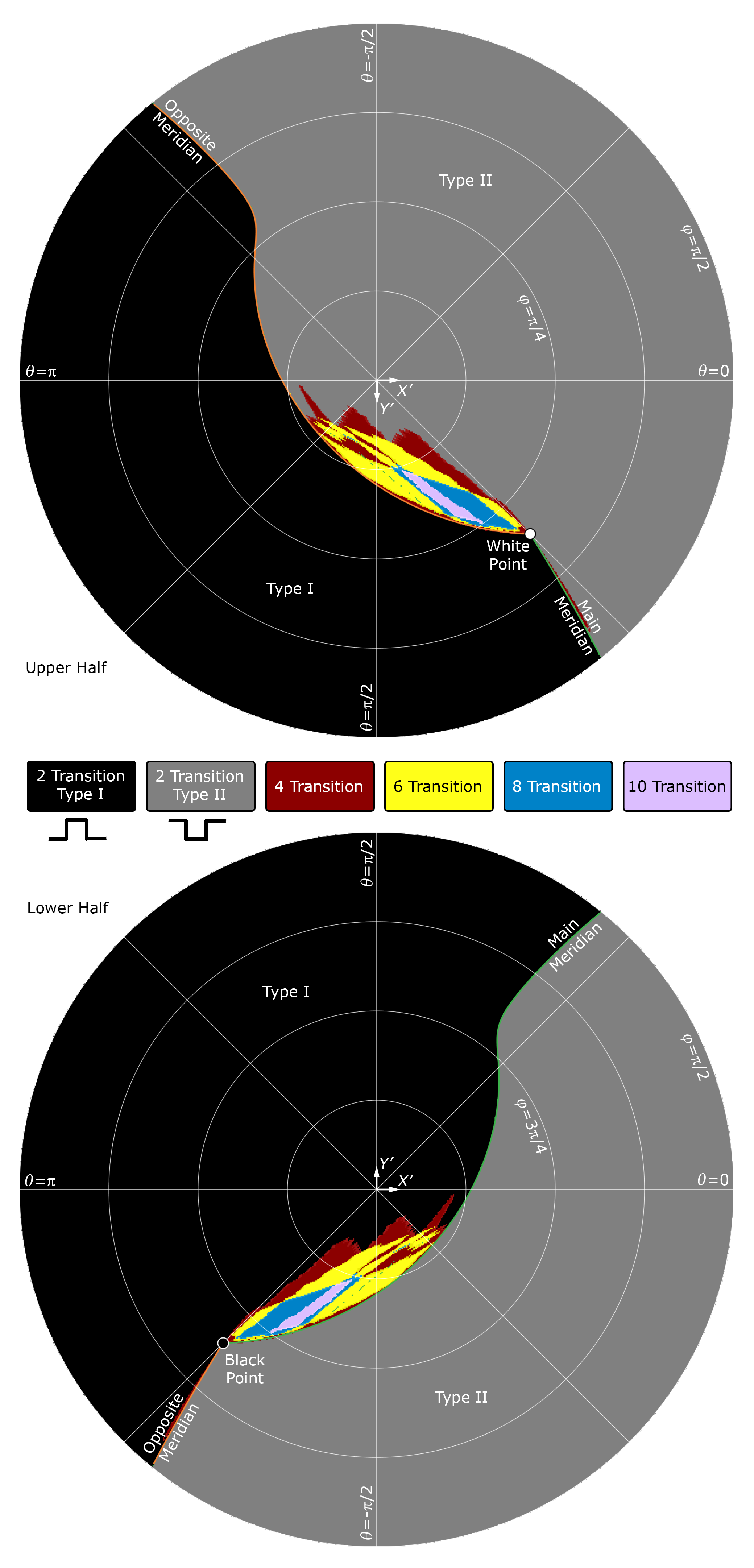}
\centering
\caption{A view of the the top and bottom halves of the OCS as viewed from the 50\% gray point, looking in the $+Z'/{-}Z'$ directions.}
\end{figure}

We are now able to create a high-resolution bitmapped graphic by superimposing a rectangular lattice over each polar plot. The lattice points correspond to the pixels of a bitmapped image, and each pixel has a corresponding spherical coordinate pair, $(\theta, \varphi)$. We can formulate a LP for each pixel by defining ${XYZ}_{targ} = {XYZ}_{\mathit{50\%}} + [\mathrm{sin}(\varphi)\;\mathrm{cos}(\theta);\;\;\mathrm{sin}(\varphi)\;\mathrm{sin}(\theta);\;\; \mathrm{cos}(\varphi)]$. Figure 6 presents a color-coded summary of transition count of the LP solution associated with each pixel's ${XYZ}_{targ}$.

Several notable features stand out. First, it is evident that the vast majority of optimal colors have the two-transition Schrödinger form (the black and gray regions). Second it appears that there are two regions of higher-transition optimal colors that are mirror images of one another. The region in the upper half are all type II-like, having reflectance of 1 everywhere except for several pockets of 0 values. Conversely, the region in the lower half are all type I-like, having 0 reflectance everywhere except for several pockets of 1 values. Another way to interpret this is that the complement of a higher-transition color is a color of the same transition count, but of the opposite type. A line connecting the two passes through ${XYZ}_{\mathit{50\%}}$ (point symmetry).

Certainly there are two-transition object colors in the same directions as the higher-transition colors in Figure 6. How do they compare? The next section examines that question.

\renewcommand*{\thefootnote}{\fnsymbol{footnote}}
%=======================================================================

\textbf{5. Are Some Schrödinger Colors Sub-Optimal?}

The short answer appears to be yes, but only because of the slight non-convexity of the 1931 $2^{\circ}$ CIE Standard Colorimetric Observer Data. To demonstrate this, another computational experiment that parallels that shown in Figure 6 was performed, this time using each ${XYZ}_{targ}$ to identify the two-transition color along the associated ray. This was accomplished using the MATLAB code that was presented in a paper by Masaoka and Berns.$^{23}$ Its original purpose was to compute ``optimal metamers" for Logvinenko's ``object color space."$^{24}$ Here, it is adapted to compute only the two-transition part.\footnote{There is a very minor bug in the Masaoka \& Berns MATLAB code. The statement in their function ``optm" $\texttt{XYZopt = sum(T(L(1):L(2),:))...}$ should be instead $\texttt{XYZopt = sum(T(L(1):L(2),:),1)...}$. The bug occurs in the very rare instance when L(1) = L(2); in that case, the original statement incorrectly returns a scalar.} 

By computing the distance from ${XYZ}_{\mathit{50\%}}$ to the tristimulus values of both the two-transition color and the high-transition color, both measured along the same ray, a comparison can be made. In all cases, the high-transition color was found to be slightly farther away from ${XYZ}_{\mathit{50\%}}$ than the two-transition color was. Figure 7 summarizes this difference graphically. Only the top half is shown, as the bottom half is again a point symmetric copy.

\begin{figure}[h]
\captionsetup{width=0.8\linewidth}
\includegraphics[width=0.8\textwidth]{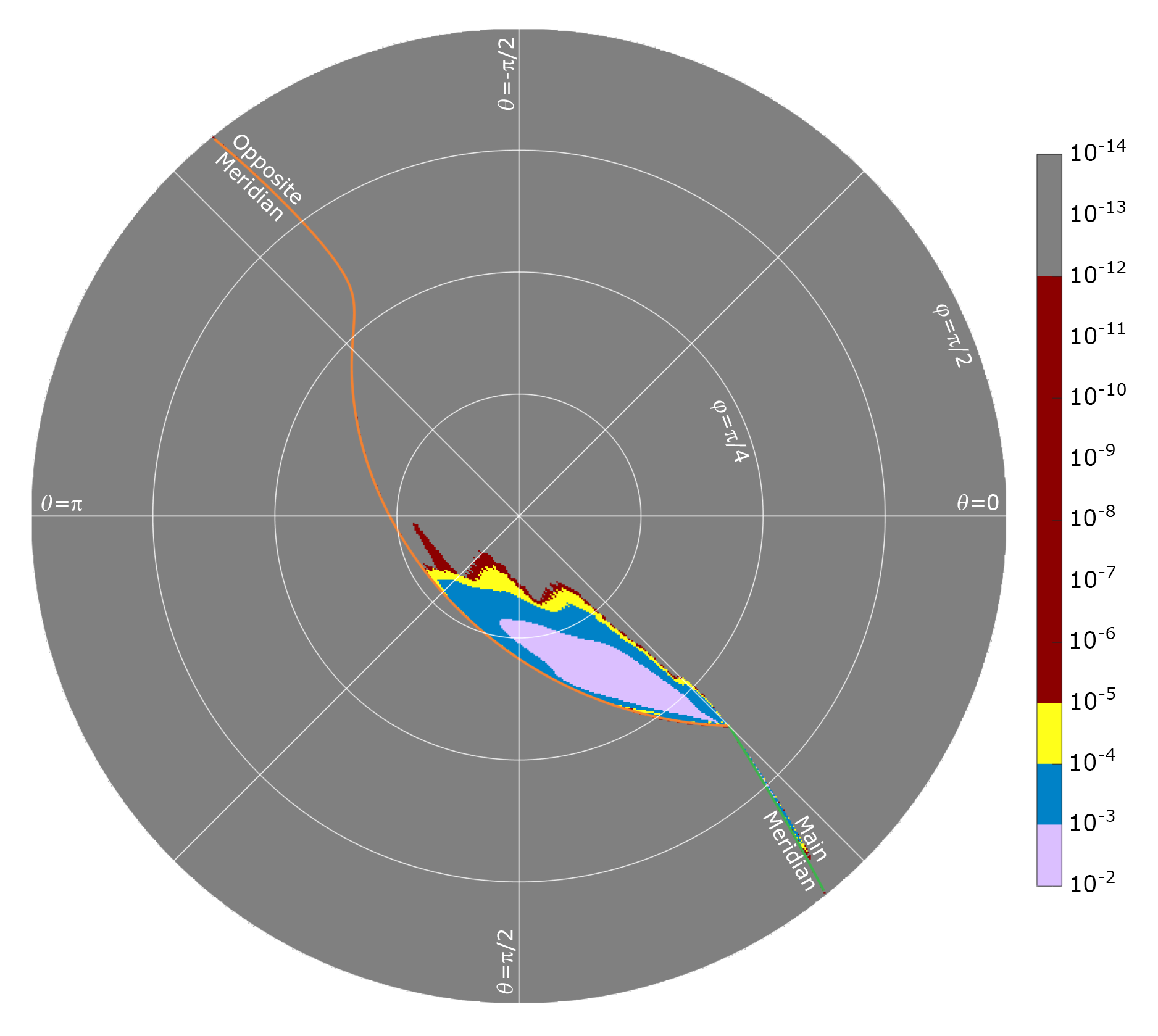}
\centering
\caption{Difference in distances to the LP solutions and to the two-transition solutions, measured from ${XYZ}_{\mathit{50\%}}$, both along the same ray.}
\end{figure}

The gray region is where the LP solution gives a two-transition color, which matches the 2-transition color returned by the Masaoka \& Berns code. In those cases, the distances from ${XYZ}_{\mathit{50\%}}$ of the two match to machine floating point tolerance. In regions of higher transition count, however, the difference in distances is ten orders of magnitude greater than machine tolerance. We can conclude that the two-transition colors in the regions where the LP gives higher-transition optimal colors are sub-optimal and fall slightly inside the OCS surface.

%=======================================================================

\textbf{6. Effect of Changing CMFs Resolution}

In Section 2, it was shown that the 1931 $2^{\circ}$ CIE CMFs retain non-convexity at other wavelength interval sizes. Here we examine one specific point on the OCS surface and how the LP high-transition optimal color and two-transition color (from the Masaoka \& Berns code) are affected by resolution of the CMFs. Figure 8 shows the two solutions for wavelength resolutions of 0.1 nm, 1 nm, 5 nm, and 10 nm. The black line is the two-transition reflectance curve and the blue region is the LP optimal reflectance curve. (Note that both are bar plots, one is filled in blue and the other is only outlined.) The small image inset into each of the four plots shows the transition count distribution as computed from the CMFs with the various wavelength interval sizes. It uses the same color coding as shown in Figure 6. Even at 10 nm intervals, the high-transition behavior persists. The small white dot in each plot shows where the two reflectance plots are located on the OCS surface.

\begin{figure}[p]
\captionsetup{width=0.73\linewidth}
\includegraphics[width=0.73\textwidth]{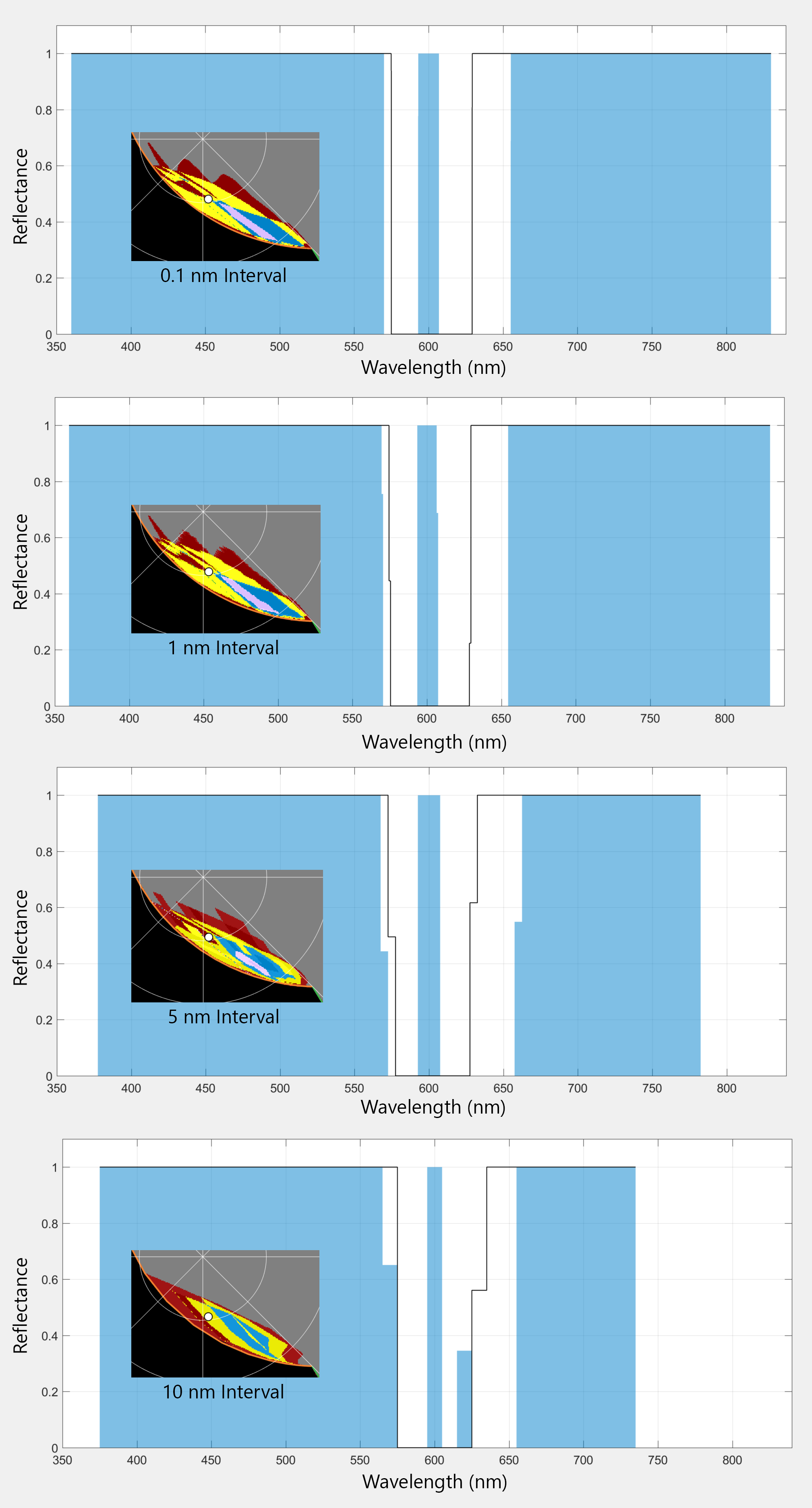}
\centering
\caption{The effect of different CMFs wavelength interval size.}
\end{figure}

The reflectance distributions all show magnitudes at the transition points that are between 0 and 1. This is an artifact of the discretization of the CMFs. Since both the LP and the Masaoka \& Berns code have constraints the require the reflectance curve to have an associated tristimulus value triplet that exactly matches a specified value, the only way to satisfy this match is to allow the transition points to have intermediate values. The way this should be interpreted is that the ``true" transition points fall somewhere within the finite width of the wavelength interval. Li et al noticed this same artifact in their 2010 paper, and came to the same conclusion.$^{22}$

As mentioned in Section 5, the difference in tristimulus values between the LP solution and the two-transition solution is very small. For example, for the 1 nm case shown in Figure 8, we have the following numerical values:

\vspace{-0.8cm}
\begin{equation}
\begin{split}
&\theta = 1.478858\;\mathrm{rad}, \;\;\varphi = 0.371322\;\mathrm{rad},\;\;{XYZ}_{targ} = \begin{Bmatrix} 50.03731 \\ 50.36132 \\ 50.94838 \end{Bmatrix}\\
&{XYZ}_{LP soln} = \begin{Bmatrix} 51.79069 \\ 69.37875 \\ 99.99523 \end{Bmatrix},\;\;\;{XYZ}_{two-trans} = \begin{Bmatrix} 51.79066 \\ 69.37828 \\ 99.99402 \end{Bmatrix}\\
&\mathrm{Difference\;in\;distance\;from}\;{XYZ}_{\mathit{50\%}} = 1.29\times 10^{-3}
\end{split}
\end{equation}

%=======================================================================

\textbf{7. Other Illuminants}

So far, all results presented are for an equal-energy illuminant. Here we examine how other illuminants affect the results. Five high-chroma Munsell colors were selected, 5R 5/14, 5Y 8/16, 5G 7/10, 5B 6/10, and 5P 4/12, and an illuminant was created with the same chromaticity, using a ``smoothest" reflectance reconstruction technique.$^{25-26}$ They are all shown at the top of Figure 9, scaled to have a maximum value of 1. 

\begin{figure}[!]
\captionsetup{width=0.83\linewidth}
\includegraphics[width=0.83\textwidth]{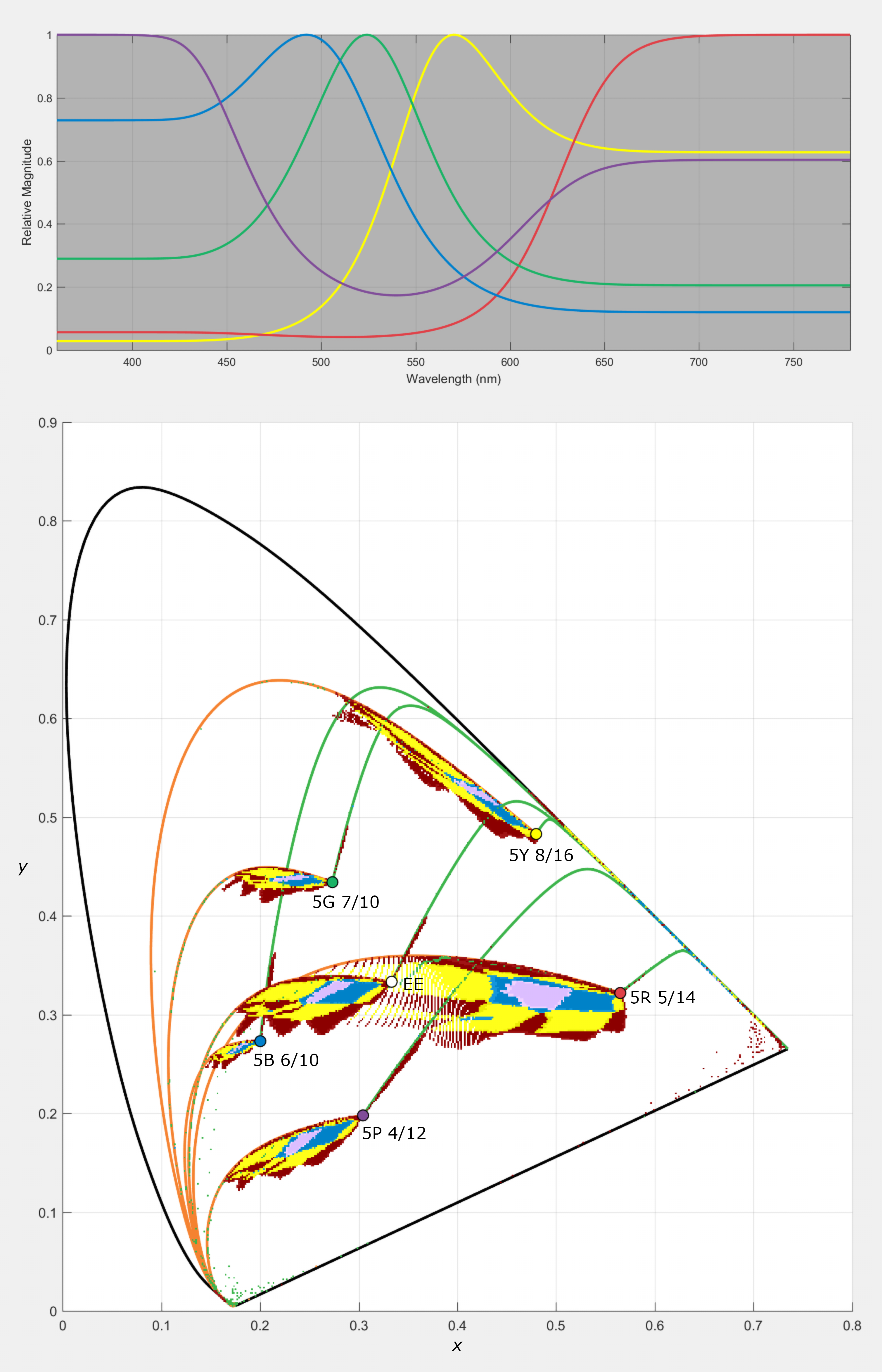}
\centering
\caption{Five illuminants with chromaticities matching Munsell 5R 5/14, 5Y 8/16, 5G 7/10, 5B 6/10, and 5P 4/12 and high-transition optimal color regions associated with each illuminant, plotted on the chromaticity diagram.}
\end{figure}

The lower portion of Figure 9 shows the regions of high-transition optimal colors associated with each of the Munsell-based illuminants, plotted on the chromaticity plane. The small circles are the chromaticities of the illuminants. The green and orange lines are the main and opposite meridians. Only the type II-like regions appear since the type I-like regions have very high chromatic purity, and are compressed along the boundary of the chromaticity diagram. It is evident that the general distribution of transition count locations persists across a wide range of illuminants. The choppiness in Figure 9 is simply an artifact of a difference in graphics resolution between Figure 6 and 9.

%=======================================================================

\textbf{8. Conclusions}

The main conclusion that can be drawn from this presentation is that very slight non-convexities of the spectral locus can have a significant impact on the shape of reflectance curves associated with optimal object colors. It is possible that the non-convexity was introduced by the manipulation of the Wright-Guild data leading to the 1931 standard.$^{27}$ The slight non-convexity has no negative impact on everyday colorimetric computations.

Another interesting observation is that the high-transition solutions appear to come in point-symmetric pairs, one type I-like and the other type II-like, but otherwise matching in transition locations. The size of the regions of high-transition optimal colors, when plotted on the chromaticity diagram, depends on the illuminant. The largest regions are associated with illuminants nearest to the boundary non-convexities.

The use of linear programming to find optimal reflectance spectra is not new, but the way of searching along an arbitrary ray through a target is unique. This brings up other potential applications. For example, to create a slice of the three-dimensional OCS for some fixed value of $X$, $Y$, or $Z$, all we need to do is relocate our coordinate system to the point on the gray line (connecting the origin and the white point) that has the desired value of $X$, $Y$, or $Z$, and then sweep the search ray around a circle in the plane of the other two tristimulus coordinates.

Additional supporting material and additional insights can be found at the online supplementary documentation for this paper.$^{27}$

\textbf{References}

\setstretch{1.0}

[1] Kuehni RG. Erwin Schrödinger, Theorie der Pigmente von grösster Leuchtkraft (Theory of pigments of greatest lightness). http://www.iscc-archive.org/pdf/SchroePigments2.pdf Also: 
https://web.archive.org/web/20190120185844/http://www.iscc-archive.org/pdf/SchroePigments2.pdf  Short URL: https://bit.ly/39fUdkv Accessed March 26, 2021.

[2] MacAdam DL. The theory of the maximum visual efficiency of colored materials. \textit{J Opt Soc Am.} 1935;25(8):249–252.

[3] MacAdam DL. Maximum visual efficiency of of colored materials. \textit{J Opt Soc Am.} 1935;25(11):361–367.

[4] Wyszecki G, Stiles WS. \textit{Color Science, Concepts and Methods, Quantitative Data and Formulae.} 2nd ed. New York, NY: John Wiley \& Sons; 1982:179-184.

[5] MacAdam DL. \textit{Color Measurement, Theme and Variations.} 2nd ed. Berlin Heidelberg New York: Springer-Verlag; 1985:122.

[6] Koenderink JJ., van Doorn AJ. Perspectives on colour space. In: Mausfeld R, Heyer D eds. \textit{Colour Vision: From Light to Object.} Oxford, UK: Oxford University Press; 2003:1–56.

[7] Kuehni RG. \textit{Color, An Introduction to Practice and Principles.} 3rd ed. Hoboken, NJ: John Wiley \& Sons; 2013:126.

[8] Oleari C. \textit{Standard Colorimetry, Definitions, Algorithms and Software.} Chichester, West Sussex, UK: John Wiley \& Sons; 2016:352.

[9] West G, Brill M. Conditions under which Schrödinger object colors are optimal. \textit{J Opt Soc Am.} 1983;73(9):1223-1225.

[10] Davis G. Convexity and Transitions, a strict examination of the 1931 CIE inverted-U, April 1, 2020. https://cran.r-project.org/web/packages/colorSpec/vignettes/convexity.pdf. Accessed March 26, 2021.

[11] University College London, Color \& Vision Research Laboratory. Color Matching Functions. http://cvrl.ioo.ucl.ac.uk/cmfs.htm. Accessed March 3, 2021.

[12] Rochester Institute of Technology, Munsell Color Science Laboratory. Useful color data.\\
https://www.rit.edu/cos/colorscience/rc\_useful\_data.php Short URL: 
https://bit.ly/39hBp4e Excel data at http://www.rit-mcsl.org/UsefulData/all\_1nm\_data.xls Short URL: https://bit.ly/3srkdAL Accessed March 26, 2021.

[13] Martínez-Verdú F, Perales E, Chorro E, de Fez D, Viqueira V, Gilabert E. Computation and visualization of the MacAdam limits for any lightness,
hue angle, and light source. \textit{J Opt Soc Am A.} 2007;24(6):1501-1515.

[14] Allen E, Some new advances in the study of metamerism: Theoretical limits of metamerism; An index of metamerism for observer differences. \textit{Color Eng.} 1969;7(1):35-40.

[15] Ohta N, Wyszecki G. Theoretical chromaticity-mismatch limits of metamers viewed under different illuminants. \textit{J Opt Soc Am.} 1975;65(3):327-333.

[16] Centore P. An open-source algorithm for metamer mismatch bodies, February 4, 2017.\\
https://www.munsellcolourscienceforpainters.com/ColourSciencePapers/AnOpenSourceAlgorithm\\ForMetamerMismatchBodies.pdf Short URL: https://bit.ly/3fhUBCK Accessed March 26, 2021

[17] Mackiewicz M, Rivertz HJ, Finlayson G. Spherical sampling methods for the calculation of metamer mismatch volumes. \textit{J Opt Soc Am A.} 2019;36(1):96-104.

[18] Ohta N, Wyszecki G. Designing illuminants that render given objects in prescribed colors. \textit{J Opt Soc Am.} 1976;66(3):269-275.

[19] Allen E. Basic equations used in computer color matching. \textit{J Opt Soc Am.} 1966;56(9):1256-1259.

[20] Belanger PR. Linear-programming approach to color recipe formulations.
\textit{J Opt Soc Am.} \\1974;64(11):1541-1544.

[21] Cogno JA, Jungman D, Conno JC. Linear and quadratic optimization algorithms for computer color matching. \textit{Color Res Appl.} 1988;13(1):40-42.

[22] Li C, Luo MR, Cho MS, Kim JS. Linear programming method for computing the gamut of object color solid. \textit{J Opt Soc Am A.} 2010;27(5):985-991.

[23] Masaoka K, Berns RS. Computation of optimal metamers. \textit{Optics Letters.} 2013;38(5):754-756.

[24] Logvinenko AD. An object-color space. \textit{J Vision.} 2009;9(11):1-23.

[25] Burns SA. Numerical methods for smoothest reflectance reconstruction. \textit{Color Res Appl.} 2020;45(1):8-21.

[26] Burns SA. Chromatic adaptation transform by spectral reconstruction. \textit{Color Res Appl.} 2019;44(5):682-693.

[27] Burns SA. Supplementary Documentation: The Location of Optimal Object Colors with More Than Two Transitions.  http://scottburns.us/supplementary-documentation-the-location-of-optimal-object-colors-with-more-than-two-transitions/ Short URL: https://bit.ly/3rxEZh7 Accessed March 26, 2021.

\end{document}